\documentclass[11pt,a4paper]{article}
\usepackage[hyperref]{emnlp2020}
\usepackage{times}
\usepackage{latexsym}
\usepackage{url}
\usepackage{multirow}

\usepackage{amsmath}
\usepackage{booktabs}
\usepackage{graphicx}
\usepackage{adjustbox}
\usepackage{subcaption}

\usepackage{todonotes}  

\aclfinalcopy 

\title{Towards End-to-End In-Image Neural Machine Translation}

\author{Elman Mansimov$^{1,*}$ ~~ Mitchell Stern$^{2,*}$ \\ \textbf{Mia Chen}$^3$ ~~ \textbf{Orhan Firat}$^3$ ~~ \textbf{Jakob Uszkoreit}$^3$~~ \textbf{Puneet Jain}$^3$\\
$^1$New York University ~~ $^2$UC Berkeley ~~ $^3$Google Research\\
$^*$Equal Contribution\\
{\tt mansimov@cs.nyu.edu, mitchell@berkeley.edu} \\}

\date{}

\begin{document}

\maketitle

\begin{abstract}
In this paper, we offer a preliminary investigation into the task of in-image machine translation: transforming an image containing text in one language into an image containing the same text in another language. We propose an end-to-end neural model for this task inspired by recent approaches to neural machine translation, and demonstrate promising initial results based purely on pixel-level supervision. We then offer a quantitative and qualitative evaluation of our system outputs and discuss some common failure modes. Finally, we conclude with directions for future work.
\end{abstract}

\section{Introduction}
\vspace{-5px}

End-to-end neural models have emerged in recent years as the dominant approach to a wide variety of sequence generation tasks in natural language processing, including speech recognition, machine translation, and dialog generation, among many others. While highly accurate, these models typically operate by outputting tokens from a predetermined symbolic vocabulary, and require integration into larger pipelines for use in user-facing applications such as voice assistants where neither the input nor output modality is text.

In the speech domain, neural methods have recently been successfully applied to end-to-end speech translation \cite{jia2019direct,liu2019endtoend,inaguma2019multilingual}, in which the goal is to translate directly from speech in one language to speech in another language. We propose to study the analogous problem of in-image machine translation. Specifically, an image containing text in one language is to be transformed into an image containing the same text in another language, removing the dependency of any predetermined symbolic vocabulary or processing.

\paragraph{Why In-Image Neural Machine Translation ?} In-image neural machine translation is a compelling test-bed for both research and engineering communities for a variety of reasons. Although there are existing commercial products that address this problem such as image translation feature of Google Translate\footnote{\href{https://blog.google/products/translate/google-translates-instant-camera-translation-gets-upgrade}{blog.google/translate/instant-camera-translation}} the underlying technical solutions are unknown. By leveraging large amounts of data and compute, end-to-end neural system could potentially improve overall quality of pipelined approaches for image translation. 
Second, and arguably more importantly, working directly with pixels has the potential to sidestep issues related to vocabularies, segmentation, and tokenization, allowing for the possibility of more universal approaches to neural machine translation, by unifying input and output spaces via pixels.

Text preprocessing and vocabulary construction has been an active research area leading to work on investigating neural machine translation systems operating on subword units \citep{sennrich2016bpe}, characters \citep{lee2017character} and even bytes \citep{wang2019bytes} and has been highlighted to be one of the major challenges when dealing with many languages simultaneously in multilingual machine translation \citep{google2019multinmt}, and cross-lingual natural language understanding \cite{conneau2019unsupervised}. Pixels serve as a straightforward way to share vocabulary among all languages at the expense of being a significantly harder learning task for the underlying models.

In this work, we propose an end-to-end neural approach to in-image machine translation that combines elements from recent neural approaches to the relevant sub-tasks in an end-to-end differentiable manner. We provide the initial problem definition and demonstrate promising first qualitative results using only pixel-level supervision on the target side. We then analyze some of the errors made by our models, and in the process of doing so uncover a common deficiency that suggests a path forward for future work.

\section{Data Generation}
\vspace{-5px}

To our knowledge, there are no publicly available datasets for the task of in-image machine translation task. Since collecting aligned natural data for in-image translation would be a difficult and costly process, a more practical approach is to bootstrap by generating pairs of rendered images containing sentences from the WMT 2014 German-English parallel corpus. The dataset consists of 4.5M German-English parallel sentence pairs. We use newstest-2013 as a development set. For each sentence pair, we create a minimal web page for the source and target, then render each using Headless Chrome\footnote{\href{https://developers.google.com/web/updates/2017/04/headless-chrome}{developers.google.com/headless-chrome}} to obtain a pair of images. The text is displayed in a black 16-pixel sans-serif font on a white background inside of a fixed-size 1024x32-pixel frame. For simplicity, all sentences are vertically centered and left-aligned without any line-wrapping. The consistent position and styling of the text in our synthetic dataset represents an ideal scenario for in-image translation, serving as a good test-bed for initial attempts. Later, one could generalize to more realistic settings by varying the location, size, typeface, and perspective of the text and by using non-uniform backgrounds.

\section{Model}
\vspace{-5px}

Our goal is to build a neural model for the in-image translation task that can be trained end-to-end on example image pairs $(X^*, Y^*)$ of height and width $H$ and $W$ using only pixel-level supervision. We evaluate two approaches for this task: convolutional encoder-decoder model and full model that combines soft versions of the traditional pipeline in order to arrive at a modular yet fully differentiable solution.

\subsection{Convolutional Baseline}
\label{subsec:baseline-model}

Inspired by the success of convolutional encoder-decoder architectures for medical image segmentation \cite{ronneberger2015unet}, we begin with a U-net style convolutional baseline. In this version of the model, the source image $X^*$ is first compressed into a single continuous vector $h_{\text{enc}}$ using a convolutional encoder $h_{\text{enc}} = \text{enc}(X^*)$. Then, the compressed representation is used as the input to a convolutional decoder that aims to predict all target pixels in parallel. Decoder outputs the probabilities of each pixel $p(Y) = \prod_{i=1}^{H}\prod_{j=1}^{W} \text{softmax}(\text{dec}(h_{\text{enc}}))$. The convolutional encoder consists of four residual blocks with the dimensions shown in Table~\ref{tab:convolutional-encoder}, and the convolutional decoder uses the same network structure in reverse order, composing a simple encoder-decoder architecture with a representational bottleneck. We threshold the grayscale value of each pixel in the groundtruth output image at 0.5 to obtain a binary black-and-white target, and use a binary cross-entropy loss on the pixels of the model output as our loss function for training.

\begin{table}[h]
\centering
\begin{tabular}{ccccc}
\toprule
Dimensions & In & Out & Kernel & Stride \\
\midrule
(1024, 32) & 3 & 64 & 3 & 1 \\
(1024, 32) & 64 & 128 & 3 & 2 \\
\midrule
(512, 16) & 128 & 128 & 3 & 1 \\
(512, 16) & 128 & 256 & 3 & 2 \\
\midrule
(256, 8) & 256 & 256 & 3 & 1 \\
(256, 8) & 256 & 512 & 3 & 2 \\
\midrule
(128, 4) & 512 & 512 & 3 & 1 \\
(128, 4) & 512 & 512 & 3 & 2 \\
\bottomrule
\end{tabular}
\caption{The parameters of our convolutional encoder network. Each block contains a residual connection from the input to the output. The decoder network uses the same structure in reverse. Dimensions correspond to the size of image. In, out, kernel and stride correspond to conv layer hyperparameters.}
\label{tab:convolutional-encoder}
\end{table}

\begin{figure}[h]
\centering
\adjustbox{trim={.0\width} {.0\height} {0.5\width} {.0\height},clip}{\includegraphics[width=\textwidth]{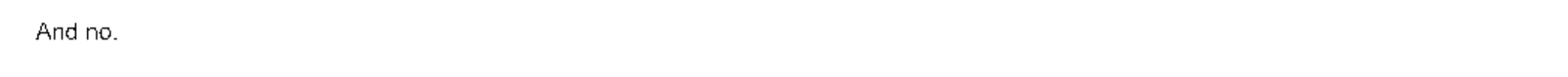}}
\adjustbox{trim={.0\width} {.0\height} {0.5\width} {.0\height},clip}{\includegraphics[width=\textwidth]{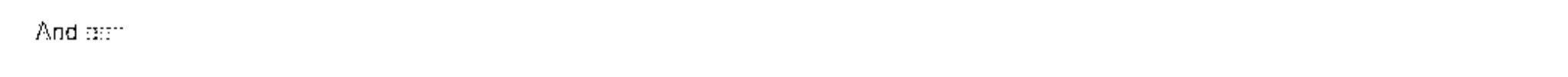}}
\adjustbox{trim={.0\width} {.0\height} {0.5\width} {.0\height},clip}{\includegraphics[width=\textwidth]{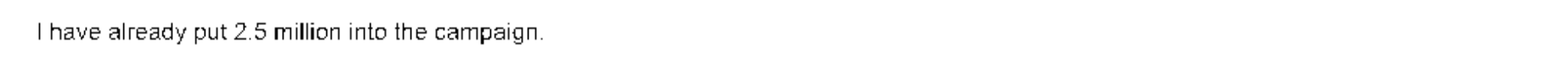}}
\adjustbox{trim={.0\width} {.0\height} {0.5\width} {.0\height},clip}{\includegraphics[width=\textwidth]{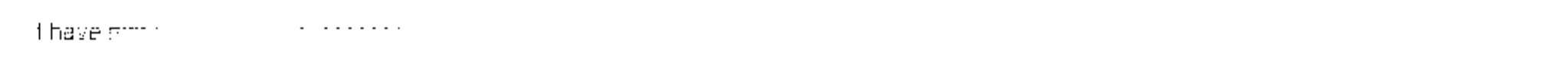}}
\caption{Example predictions made by the baseline convolutional model from Section~\ref{subsec:baseline-model}. We show two pairs of groundtruth target images followed by generated target images. Although it successfully predicts one or two words, it quickly devolves into noise thereafter.}
\label{fig:baseline-example}
\end{figure}

In order to solve the proposed task, this baseline must address the combined challenges of recognizing and rendering text at a pixel level, capturing the meaning of a sentence in a single vector as in early sequence-to-sequence models \cite{sutskever2014sequence}, and performing non-autoregressive translation \cite{gu2018nonautoregressive}. Although the model can sometimes produce the first few words of the output, it is unable to learn much beyond that; see Figure~\ref{fig:baseline-example} for a representative example.

\subsection{Full Model}

To better take advantage of the problem structure, we next propose a modular neural model that breaks the problem down into more manageable sub-tasks while still being trainable end-to-end. Intuitively, one would expect a model that can successfully carry out the in-image machine translation task to first recognize the text represented in the input image, next perform some computation over its internal representation to obtain a soft translation, and finally generate the output image through a learned rendering process. Moreover, just as modern neural machine translation systems predict the output over the span of multiple time steps in a auto-regressive way rather than all at once, it stands to reason that such a decomposition would be of use here as well.

\begin{figure}
\includegraphics[width=\linewidth]{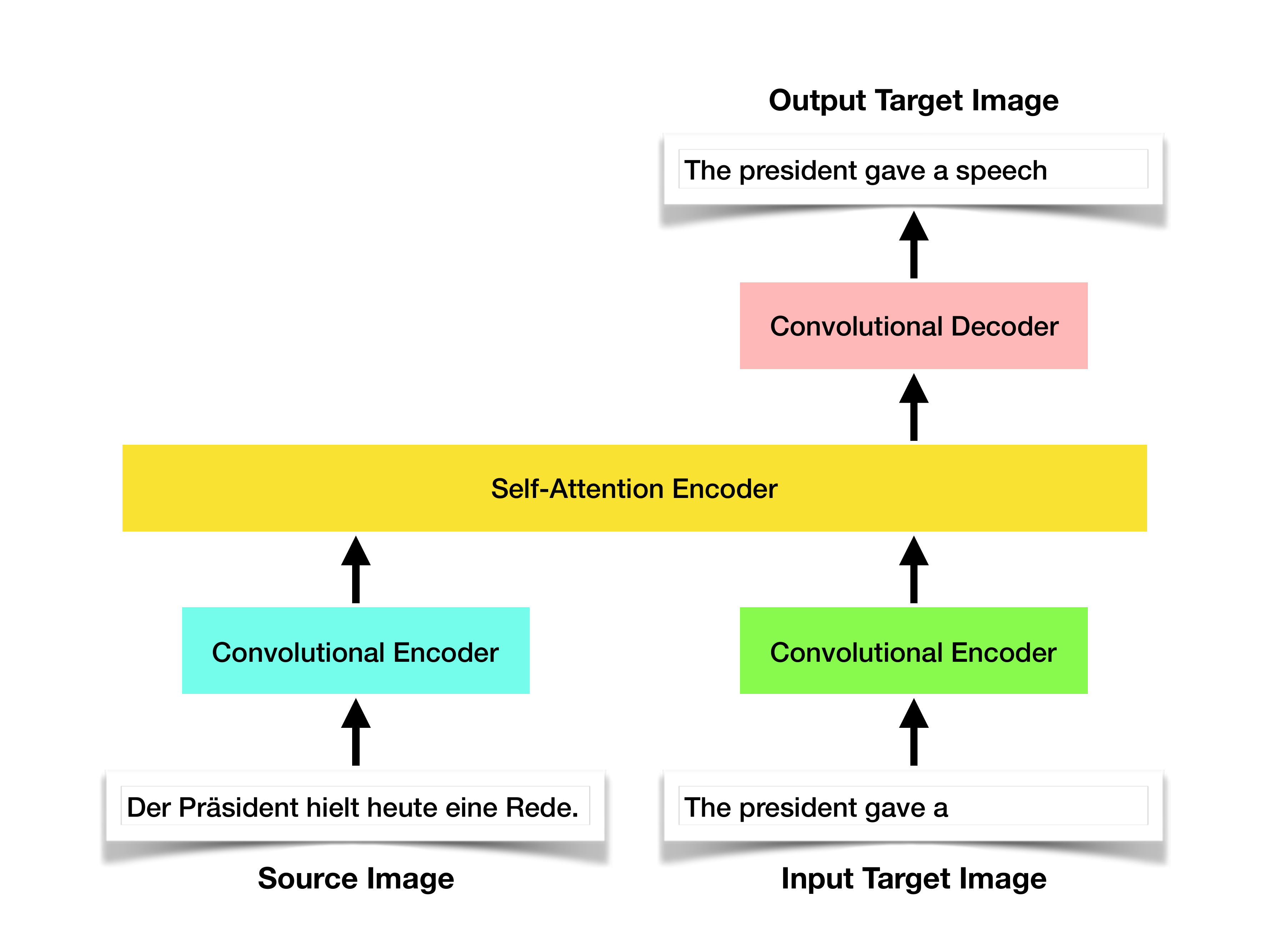}
\caption{One decoding step for our full model on an example German-English in-image translation pair. The model can be viewed as a fully differentiable analog of the more traditional OCR $\to$ translate $\to$ render pipeline.}
\label{fig:full-model}
\end{figure}

To this end, we propose a revised model that receives as input both the source image $X^{*}$ and a partial (or proposal) target image $Y_{<n}^{*}$, applies separate convolutional encoders to each source and target images in order to recognize the text contained therein. The model then applies a self-attention encoder \cite{vaswani2017attention} to the concatenated output of two convolutional encoders to extend the translation by one step, and runs the result through a convolutional decoder. The convolutional decoder is tasked to obtain a new partial output at every generation step, $Y_{\leq n}^{*}$, that is one step closer to the final target image. The model uses the same structure as the baseline for the convolutional encoder and decoder components, and includes a 6-layer self-attention encoder with hidden dimension 512 and feed-forward dimension 2048 in the middle to help carry out translation within the learned continuous representation space. A visualization of the architecture is given in Figure~\ref{fig:full-model}.

With this approach, the problem is decomposed into a sequence of image predictions, each of which conditions on the previously generated output when generating the next candidate output. We use a SentencePiece vocabulary \cite{kudo2018sentencepiece} to break the underlying sentence into sentence pieces, and decompose each example into one sub-example per sentence piece. The $n$th sub-example has a target-side input image consisting of the first $n-1$ sentence pieces, and is trained to predict an output image consisting of the first $n$ sentence pieces from the target sentence. We use the same pixel-level loss as in the baseline. Since the model fully regenerates the output at each step, it must learn to copy the portion that is already present in the target-side input in addition to predicting and rendering the next token. Decoding is done sequentially in a greedy fashion by feeding the model its own predictions (generated partial image) in place of the gold image prefixes.

The use of an external symbolic vocabulary is chosen to speed up the prototyping by making use of existing neural machine translation baselines. The sole purpose is to provide pixel spans that do not cut the characters in half, and simplify the stopping policy. A simple character-based splitting could also be used and/or a stopping policy network could be trained in exchange for increased complexity, and training/inference costs.
\vspace{-8px}

\section{Results}
\vspace{-8px}

\begin{figure*}[ht!]
\centering
\begin{subfigure}{0.49\textwidth}
\adjustbox{trim={.0\width} {.0\height} {0.45\width} {.0\height},clip}{\includegraphics[width=1.8\textwidth]{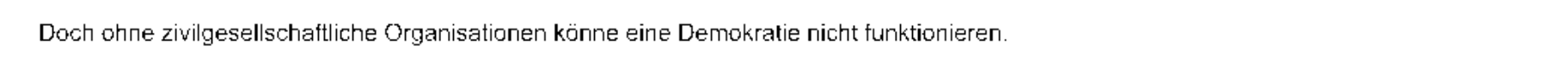}} \\
\adjustbox{trim={.0\width} {.0\height} {0.45\width} {.0\height},clip}{\includegraphics[width=1.8\textwidth]{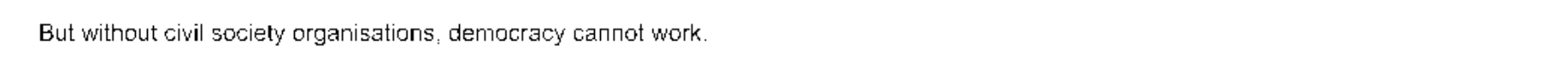}} \\
\adjustbox{trim={.0\width} {.0\height} {0.45\width} {.0\height},clip}{\includegraphics[width=1.8\textwidth]{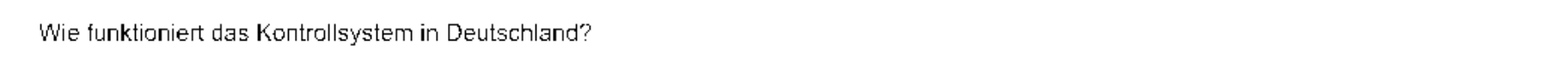}} \\
\subcaptionbox{Examples of correct predictions by our model\label{fig:full-model-preds-correct}}{\adjustbox{trim={.0\width} {.0\height} {0.45\width} {.0\height},clip}{\includegraphics[width=1.8\textwidth]{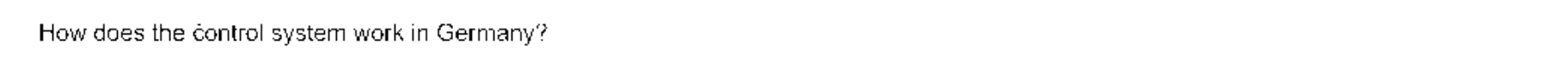}}}
\end{subfigure}
\begin{subfigure}{0.49\textwidth}
\adjustbox{trim={.0\width} {.0\height} {0.45\width} {.0\height},clip}{\includegraphics[width=1.8\textwidth]{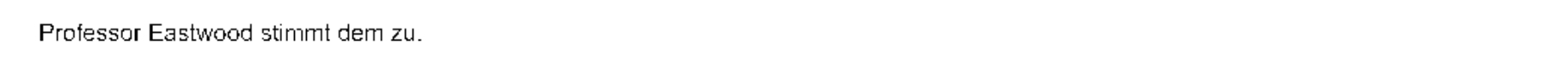}} \\
\adjustbox{trim={.0\width} {.0\height} {0.45\width} {.0\height},clip}{\includegraphics[width=1.8\textwidth]{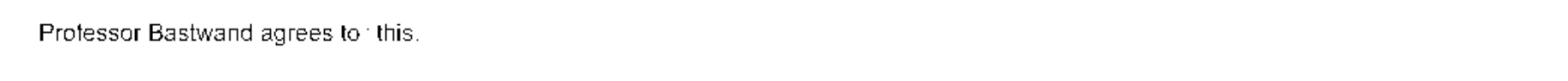}} \\
\adjustbox{trim={.0\width} {.0\height} {0.45\width} {.0\height},clip}{\includegraphics[width=1.8\textwidth]{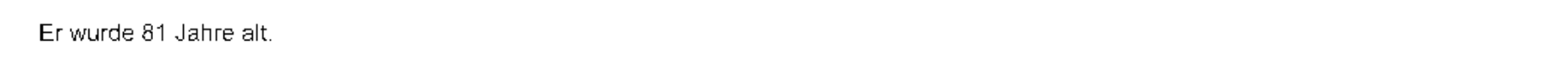}} \\
\subcaptionbox{Examples of predictions with minor typos\label{fig:full-model-preds-minor-typo}}{\adjustbox{trim={.0\width} {.0\height} {0.45\width} {.0\height},clip}{\includegraphics[width=1.8\textwidth]{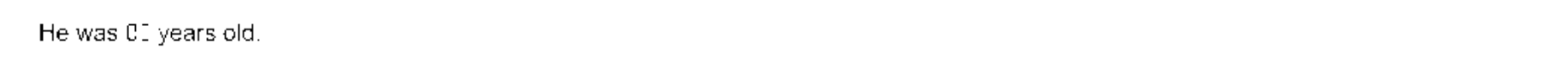}}}
\end{subfigure}

\begin{subfigure}{0.49\textwidth}
\adjustbox{trim={.0\width} {.0\height} {0.45\width} {.0\height},clip}{\includegraphics[width=1.8\textwidth]{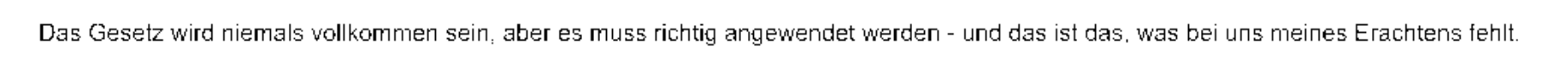}} \\
\adjustbox{trim={.0\width} {.0\height} {0.45\width} {.0\height},clip}{\includegraphics[width=1.8\textwidth]{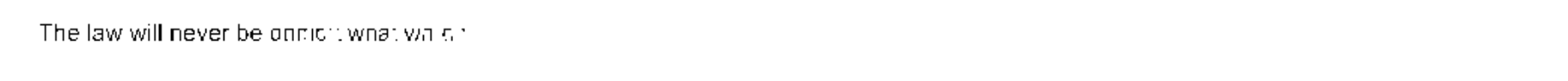}} \\
\adjustbox{trim={.0\width} {.0\height} {0.45\width} {.0\height},clip}{\includegraphics[width=1.8\textwidth]{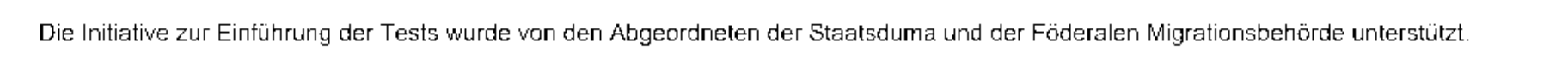}} \\
\subcaptionbox{Examples of partially correct predictions by our model\label{fig:full-model-preds-correct-start}}{\adjustbox{trim={.0\width} {.0\height} {0.45\width} {.0\height},clip}{\includegraphics[width=1.8\textwidth]{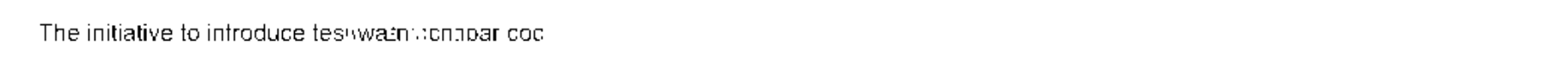}}}
\end{subfigure}
\begin{subfigure}{0.49\textwidth}
\adjustbox{trim={.0\width} {.0\height} {0.45\width} {.0\height},clip}{\includegraphics[width=1.8\textwidth]{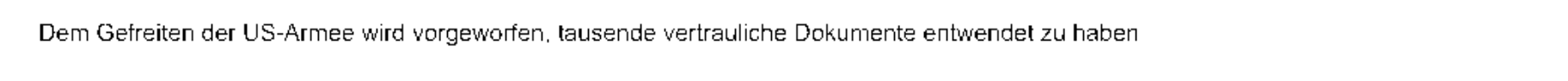}} \\
\adjustbox{trim={.0\width} {.0\height} {0.45\width} {.0\height},clip}{\includegraphics[width=1.8\textwidth]{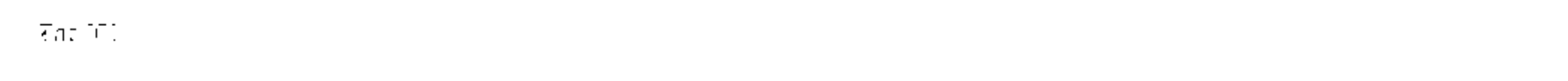}} \\
\adjustbox{trim={.0\width} {.0\height} {0.45\width} {.0\height},clip}{\includegraphics[width=1.8\textwidth]{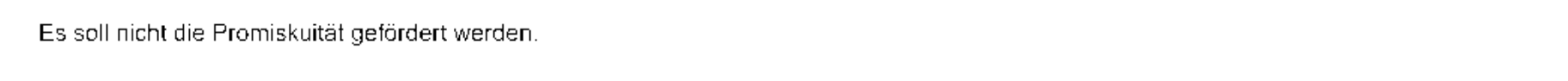}} \\
\subcaptionbox{Examples of failed predictions by our model\label{fig:full-model-preds-garbage}}{\adjustbox{trim={.0\width} {.0\height} {0.45\width} {.0\height},clip}{\includegraphics[width=1.8\textwidth]{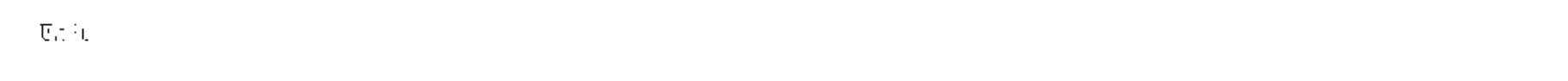}}}
\end{subfigure}
\caption{Various types of predictions made by our full model. For more qualitative results please see Appendix.}
\label{fig:full-model-preds}
\end{figure*}

\begin{figure}[ht]
\centering
\begin{subfigure}{\textwidth}
\adjustbox{trim={.0\width} {.0\height} {0.5\width} {.0\height},clip}{\includegraphics[width=\textwidth]{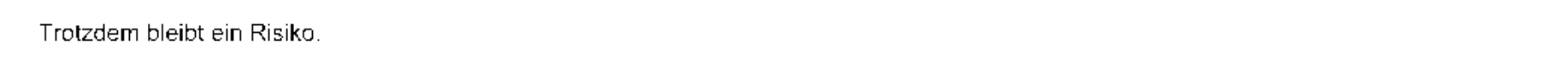}} \\
\adjustbox{trim={.0\width} {.0\height} {0.5\width} {.0\height},clip}{\includegraphics[width=\textwidth]{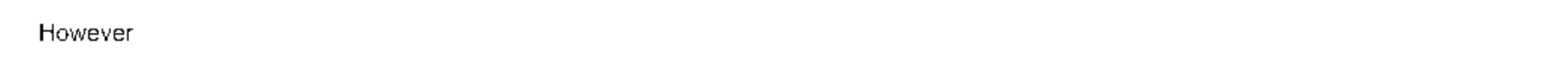}} \\
{\adjustbox{trim={.0\width} {.0\height} {0.5\width} {.0\height},clip}{\includegraphics[width=\textwidth]{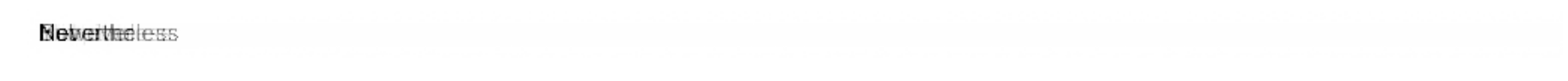}}}
\end{subfigure}
\caption{Analysis of the predictions made by the full model. First row shows the source sentence. Second row shows the groundtruth at the first timestep. Third row shows the probabilities of the pixels predicted by the full model.}
\label{fig:full-model-errors}
\end{figure}

In Figure~\ref{fig:full-model-preds} we share various source images and corresponding predicted target images generated by the full model. Despite the very high dimensionality of the output, the model occasionally succeeds at predicting the full output translation as shown in Figure~\ref{fig:full-model-preds-correct}. Additionally, Figure~\ref{fig:full-model-preds-minor-typo} shows examples where model makes a minor typo in the earlier proposals and is able to correct itself further during generation process.

Figure~\ref{fig:full-model-preds-correct-start} and Figure~\ref{fig:full-model-preds-garbage} show two major failure modes of our model where it is either able to generate the first part of the sentence and fails to generate the rest or completely fails at generating the image. To better understand the source of errors made by the full model we visualize the probabilities of pixels in Figure~\ref{fig:full-model-errors}. We can see that model is uncertain between producing word \textit{However} and \textit{Nevertheless} which leads to artifacts when taking the $\arg\max$ value of each pixel during decoding similar to ones displayed in Figure~\ref{fig:full-model-preds-garbage}.

On Table~\ref{tab:pixel-ce-and-bleu} we provide quantitative results of the both models proposed. The full model achieves significantly lower negative log likelihood score compared to convolutional baseline due to an easier task of predicting parts of the image. Neither of the models overfit on the development set. We further quantitatively measure our models by transcribing the generated images into text with a neural OCR model and measuring the BLEU \citep{papineni2002bleu} score. Convolutional baseline fails to produce images that contain transcribed text and is significantly outperformed by our full model in terms of BLEU score. 

\begin{table}[t]
\begin{center}
\begin{tabular}[b]{ccc}
\toprule
Model & Train Set & Dev Set \\
\midrule
\multirow{2}{*}{\rotatebox[origin=c]{90}{\scriptsize NLL}}
Conv Baseline & 0.120 & 0.117 \\
Full Model & 0.028 & \textbf{0.025} \\
\midrule
\multirow{2}{*}{\rotatebox[origin=c]{90}{\scriptsize BLEU}}
Conv Baseline & - & 0.5 \\
Full Model & - & \textbf{7.7} \\
\bottomrule
\end{tabular}
\end{center}
\caption{Per pixel loss function (NLL$\downarrow$) and generation quality (BLEU$\uparrow$) of convolutional baseline and full model on WMT'14 German-English Train/Dev sets.}
\label{tab:pixel-ce-and-bleu}
\end{table}

\section{Conclusion}
\vspace{-10px}
In this paper, we introduce the task of in-image neural machine translation and develop an end-to-end model that shows promising results on learning to translate text through purely pixel-level supervision. By doing so, we demonstrate a viable first step towards applying such models in more natural settings, such as translating texts, menus, or street signs within real-world images. Future work should explore models that do not rely on off-the-shelf text tokenizers to decompose the very hard image generation problem into sequence of simpler image predictions. We hypothesize that discrete latent variables \citep{vandenoord2017discretelatents} are best suited to implicitly segment the image and capture the sequential nature of this task.





\bibliographystyle{acl_natbib}
\bibliography{main}

\newpage

\section{Appendix}
\label{sec:appendix}

\begin{figure}[h]
\centering
\adjustbox{trim={.0\width} {.0\height} {0.5\width} {.0\height},clip}{\includegraphics[width=\textwidth]{examples/correct/1-input}}
\adjustbox{trim={.0\width} {.0\height} {0.5\width} {.0\height},clip}{\includegraphics[width=\textwidth]{examples/correct/1-output}}
\adjustbox{trim={.0\width} {.0\height} {0.5\width} {.0\height},clip}{\includegraphics[width=\textwidth]{examples/correct/2-input}}
\adjustbox{trim={.0\width} {.0\height} {0.5\width} {.0\height},clip}{\includegraphics[width=\textwidth]{examples/correct/2-output}}
\adjustbox{trim={.0\width} {.0\height} {0.5\width} {.0\height},clip}{\includegraphics[width=\textwidth]{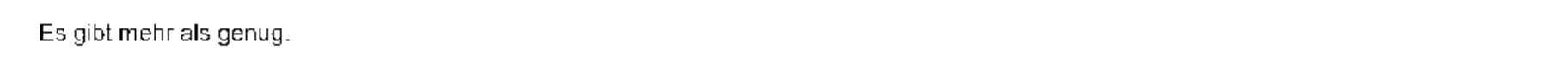}}
\adjustbox{trim={.0\width} {.0\height} {0.5\width} {.0\height},clip}{\includegraphics[width=\textwidth]{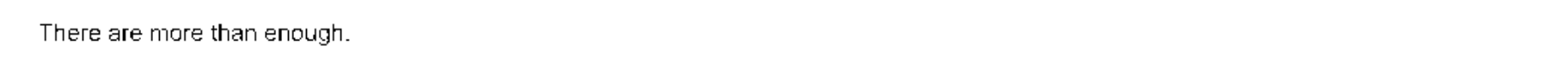}}
\adjustbox{trim={.0\width} {.0\height} {0.5\width} {.0\height},clip}{\includegraphics[width=\textwidth]{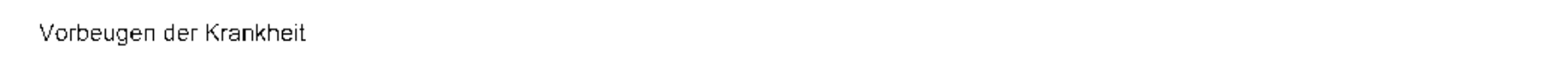}}
\adjustbox{trim={.0\width} {.0\height} {0.5\width} {.0\height},clip}{\includegraphics[width=\textwidth]{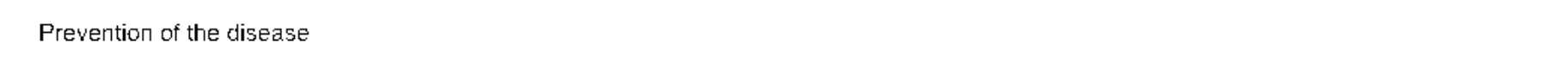}}
\adjustbox{trim={.0\width} {.0\height} {0.5\width} {.0\height},clip}{\includegraphics[width=\textwidth]{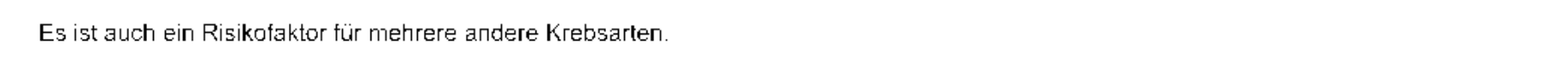}}
\adjustbox{trim={.0\width} {.0\height} {0.5\width} {.0\height},clip}{\includegraphics[width=\textwidth]{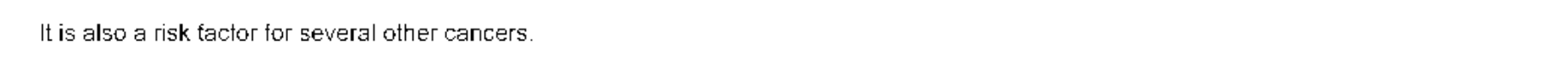}}
\adjustbox{trim={.0\width} {.0\height} {0.5\width} {.0\height},clip}{\includegraphics[width=\textwidth]{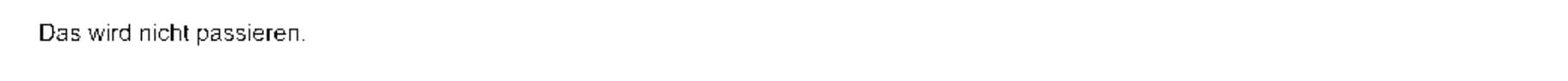}}
\adjustbox{trim={.0\width} {.0\height} {0.5\width} {.0\height},clip}{\includegraphics[width=\textwidth]{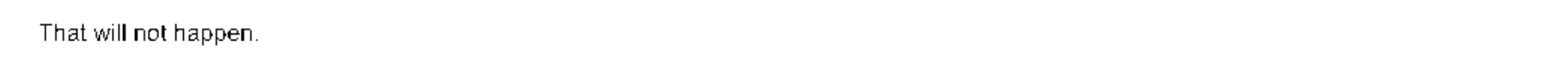}}
\adjustbox{trim={.0\width} {.0\height} {0.5\width} {.0\height},clip}{\includegraphics[width=\textwidth]{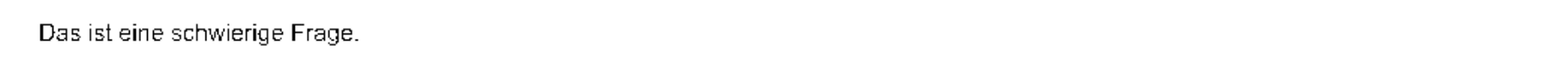}}
\adjustbox{trim={.0\width} {.0\height} {0.5\width} {.0\height},clip}{\includegraphics[width=\textwidth]{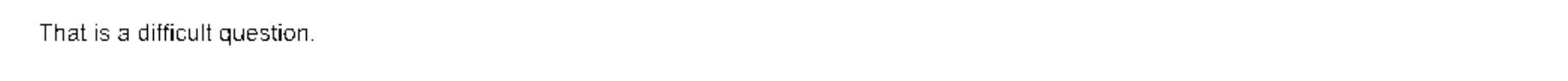}}
\caption{Example \textit{correct} predictions made by the full model on De$\rightarrow$En image translation task.}
\label{fig:appendix-full-model-correct}
\end{figure}

\newpage

\begin{figure}[h]
\centering
\adjustbox{trim={.0\width} {.0\height} {0.5\width} {.0\height},clip}{\includegraphics[width=\textwidth]{examples/minor-typo/1-input}}
\adjustbox{trim={.0\width} {.0\height} {0.5\width} {.0\height},clip}{\includegraphics[width=\textwidth]{examples/minor-typo/1-output}}
\adjustbox{trim={.0\width} {.0\height} {0.5\width} {.0\height},clip}{\includegraphics[width=\textwidth]{examples/minor-typo/2-input}}
\adjustbox{trim={.0\width} {.0\height} {0.5\width} {.0\height},clip}{\includegraphics[width=\textwidth]{examples/minor-typo/2-output}}
\adjustbox{trim={.0\width} {.0\height} {0.5\width} {.0\height},clip}{\includegraphics[width=\textwidth]{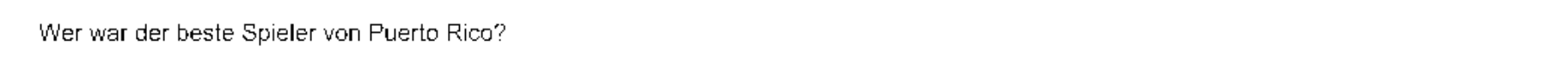}}
\adjustbox{trim={.0\width} {.0\height} {0.5\width} {.0\height},clip}{\includegraphics[width=\textwidth]{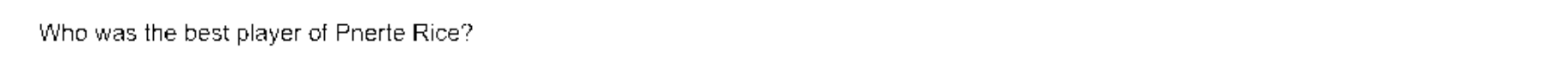}}
\adjustbox{trim={.0\width} {.0\height} {0.5\width} {.0\height},clip}{\includegraphics[width=\textwidth]{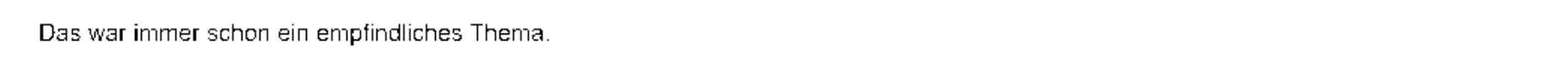}}
\adjustbox{trim={.0\width} {.0\height} {0.5\width} {.0\height},clip}{\includegraphics[width=\textwidth]{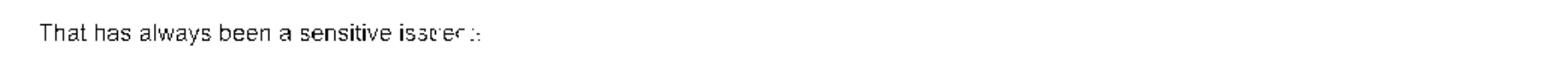}}
\adjustbox{trim={.0\width} {.0\height} {0.5\width} {.0\height},clip}{\includegraphics[width=\textwidth]{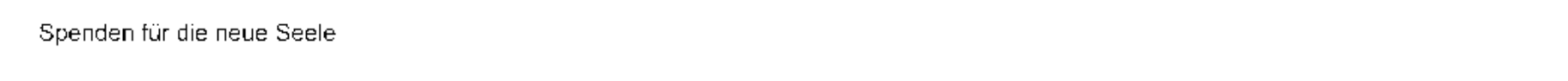}}
\adjustbox{trim={.0\width} {.0\height} {0.5\width} {.0\height},clip}{\includegraphics[width=\textwidth]{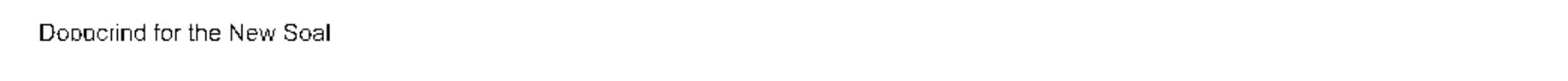}}
\adjustbox{trim={.0\width} {.0\height} {0.5\width} {.0\height},clip}{\includegraphics[width=\textwidth]{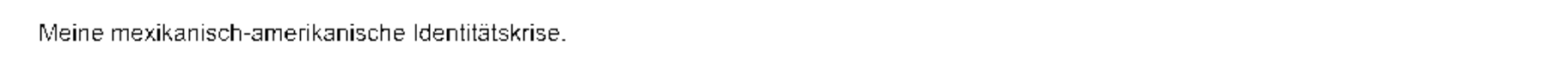}}
\adjustbox{trim={.0\width} {.0\height} {0.5\width} {.0\height},clip}{\includegraphics[width=\textwidth]{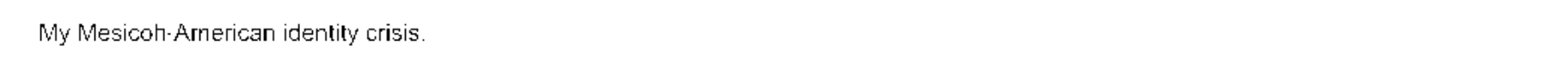}}
\adjustbox{trim={.0\width} {.0\height} {0.5\width} {.0\height},clip}{\includegraphics[width=\textwidth]{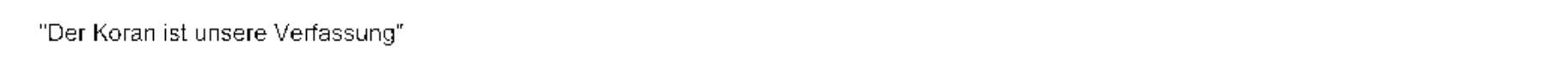}}
\adjustbox{trim={.0\width} {.0\height} {0.5\width} {.0\height},clip}{\includegraphics[width=\textwidth]{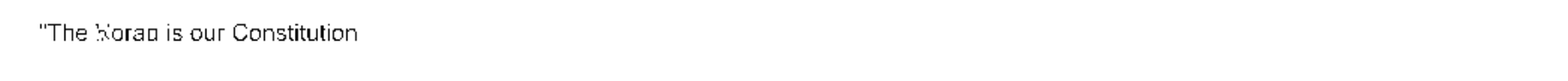}}
\caption{Example predictions with \textit{minor typo} made by the full model on De$\rightarrow$En image translation task.}
\label{fig:appendix-full-model-minor-typo}
\end{figure}

\newpage
\begin{figure}[h]
\centering
\adjustbox{trim={.0\width} {.0\height} {0.5\width} {.0\height},clip}{\includegraphics[width=\textwidth]{examples/correct-start/1-input}}
\adjustbox{trim={.0\width} {.0\height} {0.5\width} {.0\height},clip}{\includegraphics[width=\textwidth]{examples/correct-start/1-output}}
\adjustbox{trim={.0\width} {.0\height} {0.5\width} {.0\height},clip}{\includegraphics[width=\textwidth]{examples/correct-start/2-input}}
\adjustbox{trim={.0\width} {.0\height} {0.5\width} {.0\height},clip}{\includegraphics[width=\textwidth]{examples/correct-start/2-output}}
\adjustbox{trim={.0\width} {.0\height} {0.5\width} {.0\height},clip}{\includegraphics[width=\textwidth]{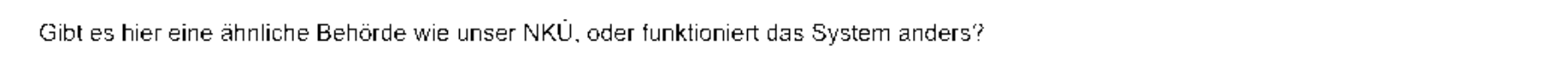}}
\adjustbox{trim={.0\width} {.0\height} {0.5\width} {.0\height},clip}{\includegraphics[width=\textwidth]{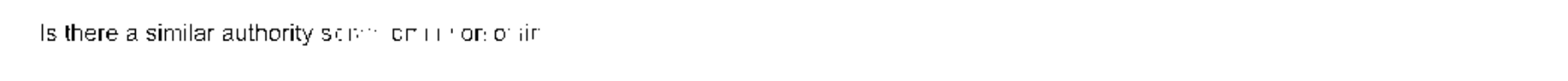}}
\adjustbox{trim={.0\width} {.0\height} {0.5\width} {.0\height},clip}{\includegraphics[width=\textwidth]{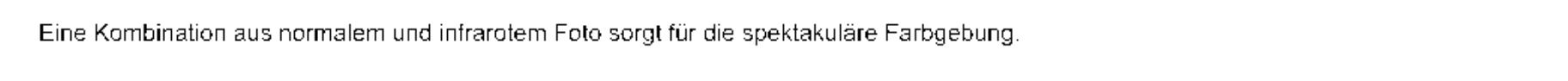}}
\adjustbox{trim={.0\width} {.0\height} {0.5\width} {.0\height},clip}{\includegraphics[width=\textwidth]{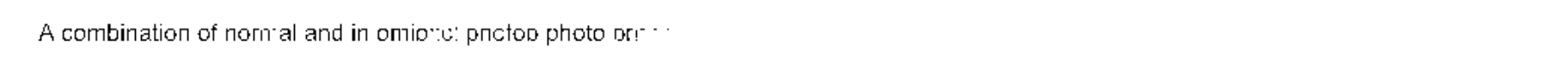}}
\adjustbox{trim={.0\width} {.0\height} {0.5\width} {.0\height},clip}{\includegraphics[width=\textwidth]{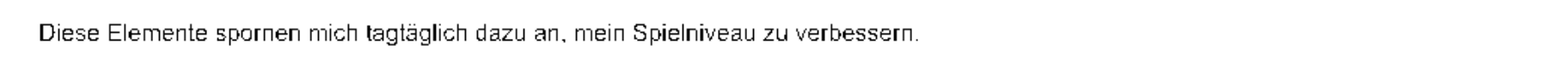}}
\adjustbox{trim={.0\width} {.0\height} {0.5\width} {.0\height},clip}{\includegraphics[width=\textwidth]{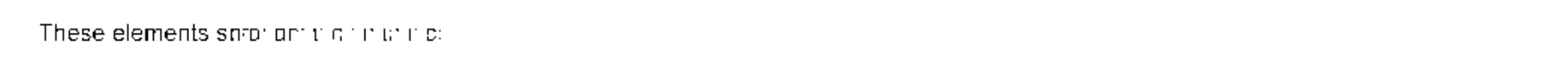}}
\adjustbox{trim={.0\width} {.0\height} {0.5\width} {.0\height},clip}{\includegraphics[width=\textwidth]{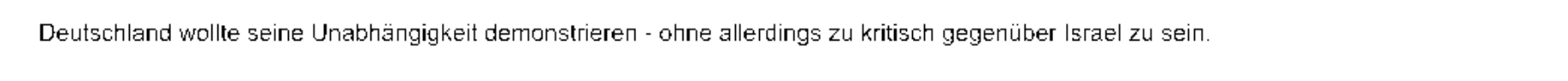}}
\adjustbox{trim={.0\width} {.0\height} {0.5\width} {.0\height},clip}{\includegraphics[width=\textwidth]{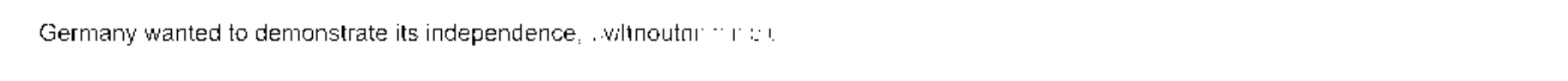}}
\adjustbox{trim={.0\width} {.0\height} {0.5\width} {.0\height},clip}{\includegraphics[width=\textwidth]{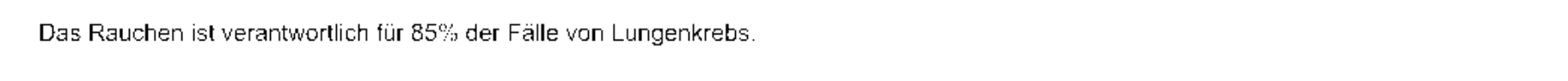}}
\adjustbox{trim={.0\width} {.0\height} {0.5\width} {.0\height},clip}{\includegraphics[width=\textwidth]{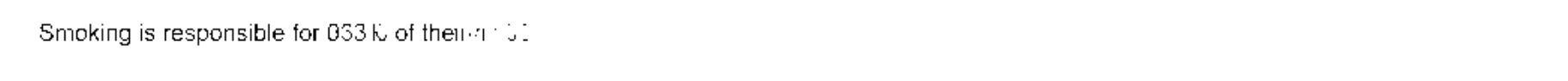}}
\caption{Example predictions with \textit{correct first few words} made by the full model on De$\rightarrow$En image translation task.}
\label{fig:appendix-full-model-correct-first}
\end{figure}

\newpage
\begin{figure}[h]
\centering
\adjustbox{trim={.0\width} {.0\height} {0.5\width} {.0\height},clip}{\includegraphics[width=\textwidth]{examples/garbage/1-input}}
\adjustbox{trim={.0\width} {.0\height} {0.5\width} {.0\height},clip}{\includegraphics[width=\textwidth]{examples/garbage/1-output}}
\adjustbox{trim={.0\width} {.0\height} {0.5\width} {.0\height},clip}{\includegraphics[width=\textwidth]{examples/garbage/2-input}}
\adjustbox{trim={.0\width} {.0\height} {0.5\width} {.0\height},clip}{\includegraphics[width=\textwidth]{examples/garbage/2-output}}
\adjustbox{trim={.0\width} {.0\height} {0.5\width} {.0\height},clip}{\includegraphics[width=\textwidth]{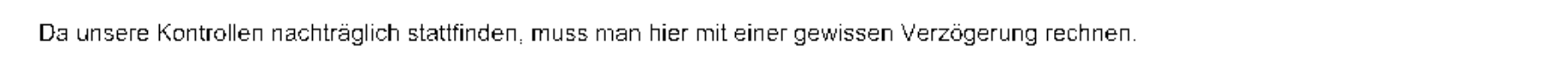}}
\adjustbox{trim={.0\width} {.0\height} {0.5\width} {.0\height},clip}{\includegraphics[width=\textwidth]{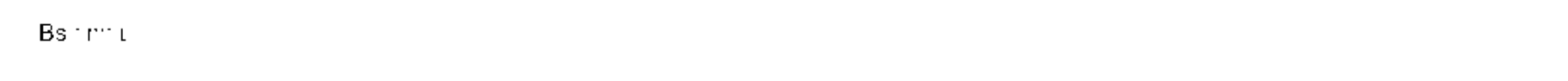}}
\adjustbox{trim={.0\width} {.0\height} {0.5\width} {.0\height},clip}{\includegraphics[width=\textwidth]{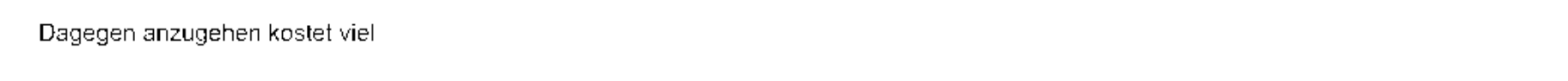}}
\adjustbox{trim={.0\width} {.0\height} {0.5\width} {.0\height},clip}{\includegraphics[width=\textwidth]{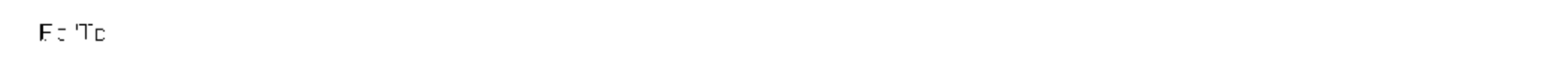}}
\adjustbox{trim={.0\width} {.0\height} {0.5\width} {.0\height},clip}{\includegraphics[width=\textwidth]{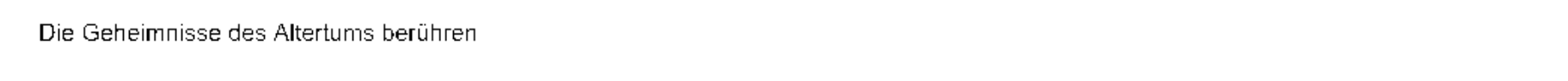}}
\adjustbox{trim={.0\width} {.0\height} {0.5\width} {.0\height},clip}{\includegraphics[width=\textwidth]{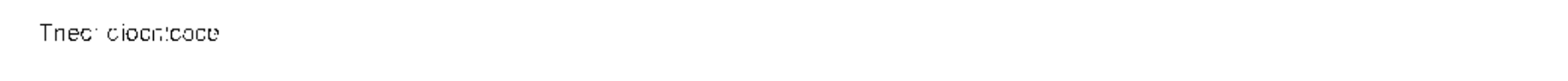}}
\adjustbox{trim={.0\width} {.0\height} {0.5\width} {.0\height},clip}{\includegraphics[width=\textwidth]{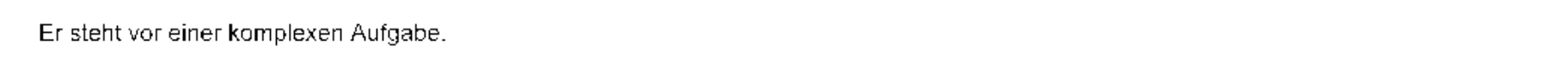}}
\adjustbox{trim={.0\width} {.0\height} {0.5\width} {.0\height},clip}{\includegraphics[width=\textwidth]{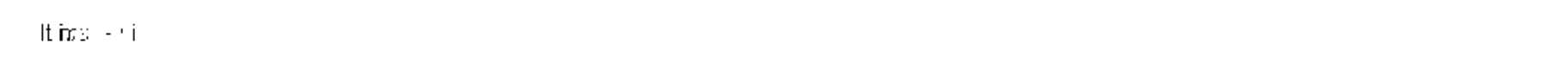}}
\caption{Example \textit{failure} predictions made by the full model on De$\rightarrow$En image translation task.}
\label{fig:appendix-full-model-garbage}
\end{figure}

\end{document}